\documentclass{article}

\usepackage[utf8]{inputenc}
\usepackage{pgfplots}
\pgfplotsset{compat=1.14}
\usepackage{tikz}
    \usetikzlibrary{shapes,decorations}
    \usetikzlibrary{positioning}
\usepackage{algorithm,algorithmic}
\usepackage{wrapfig} 
\usepackage{amsmath, amsthm, amssymb, mathrsfs, mathtools}
\usepackage{enumitem}
\usepackage{authblk}
\usepackage{hyperref}
\usepackage[numbers]{natbib}
\usepackage[capitalize]{cleveref} 

\newcommand{\cO}{\mathcal{O}}

\newcommand{\E}{\mathbb{E}}

\newcommand{\lsb}[1]{\left[#1\right]}

\newcommand{\babs}[1]{\bigl\lvert#1\bigr\rvert}

\DeclareMathOperator*{\argmax}{argmax}

\newcommand{\mypapertitle}{An Application of Online Learning to Spacecraft Memory Dump Optimization}
\newcommand{\ftl}{Follow the Leader}

\newcommand{\cOa}{\cO_{\mathrm{AOS}}}
\newcommand{\cOl}{\cO_{\mathrm{LOS}}}
\def\centerarc[#1](#2)(#3:#4:#5)
{ \draw[#1] ($(#2)+({#5*cos(#3)},{#5*sin(#3)})$) arc (#3:#4:#5); }

\theoremstyle{plain}
\newtheorem{theorem}{Theorem}[section]

\theoremstyle{definition}

\theoremstyle{remark}

\begin{document}

\title{\mypapertitle}

\author[1,2]{\textbf{Tommaso Cesari}$^*$}
\author[3,4]{\textbf{Jonathan Pergoli}$^*$}
\author[4]{\par Michele Maestrini}
\author[4]{Pierluigi Di Lizia}
\affil[1]{Universit\`a degli Studi di Milano}
\affil[2]{Toulouse School of Economics}
\affil[3]{CLC Space GmbH}
\affil[4]{Politecnico di Milano}

\maketitle

\def\thefootnote{*}\footnotetext{These authors contributed equally to this work.}\def\thefootnote{\arabic{footnote}}

\begin{abstract}
In this paper, we present a real-world application of online learning with expert advice to the field of Space Operations, testing our theory on real-life data coming from the Copernicus Sentinel-6 satellite. 
We show that in Spacecraft Memory Dump Optimization, a lightweight Follow-The-Leader algorithm leads to an increase in performance of over $60\%$ when compared to traditional techniques.
\end{abstract}

\begin{figure}
  \includegraphics[width=\textwidth]{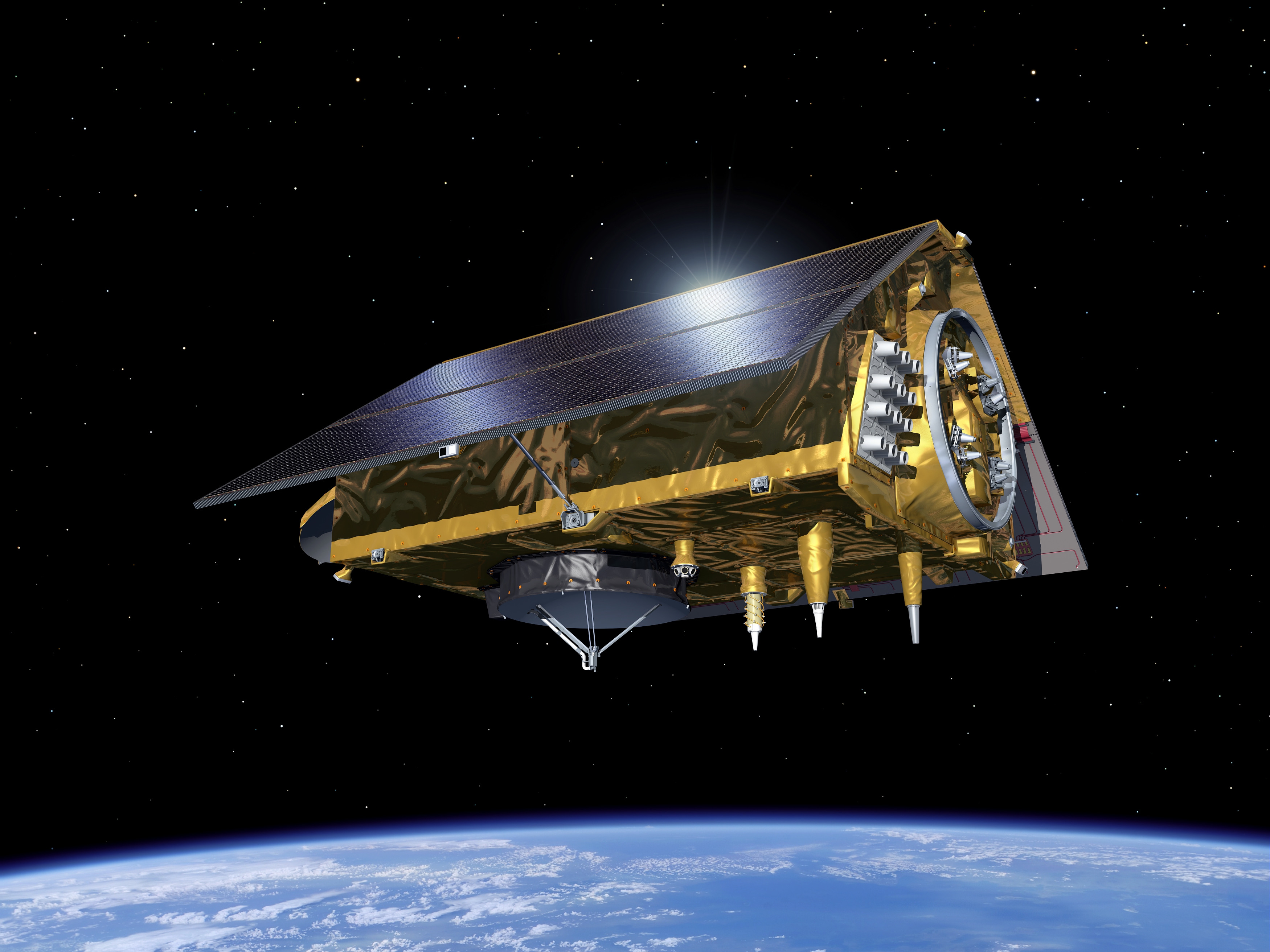}
  \caption{Copernicus Sentinel-6 satellite.}
  \label{fig:teaser}
\end{figure}

\section{Introduction}
With the fast-growing number of satellites orbiting Earth, the Space Operations field has become a prominent and thriving sector. 
As a consequence, the complexity of planning satellite operations is constantly increasing: Ground Stations have to handle communication with multiple satellites simultaneously while frequently engaged in \emph{Launch and Early Orbit Phase} (LEOP) activities; Satellite Operators need to perform routine tasks and promptly react to contingencies while checking the status of the incoming and disseminated satellite's products. 
These actions are costly, require time, and are remarkably prone to human errors.
Despite this, Satellite Operators still carry out many of these duties by relying on their technical expertise rather than leveraging modern machine learning tools.

On the other hand, computers, hardware, and flight software are becoming more sophisticated with each passing day. 
At the same time, machine learning is blooming, and optimal algorithms are constantly being designed for a variety of heterogeneous problems spanning from foundational \cite{cesa2021nonstochastic,cesari2021sample}, optimization \cite{bouttier2020regret,bachoc2021sample}, cooperation \cite{cesa2020cooperative,della2021efficient,cesa2021cooperative}, or even economics \cite{cesa2019dynamic,cesa2021aRegret,cesa2021bilateral,cesa2021roi}. 
Exploiting this growing technology means enhancing the reliability and efficiency of space operations while simultaneously avoiding human error as well as decreasing the workload on Engineers and Operators. From an operational point of view, this entails transferring Ground Segment capabilities to the Space Segment. 
As outlined by \citet{7976296}, onboard autonomy implies:

\begin{enumerate}
	\item Intelligent sensing: the capability to infer the system's state from environmental sensor data.
	\item Mission Planning and execution: the process of decomposing high-level goals into tasks that satisfy temporal and resource constraints and dispatching these tasks while monitoring and being able to react to a contingency.
	\item Fault management: the ability to deal with anomalies that occurred inside the system.
	\item Distributed decision making: effective cooperation among independent autonomous spacecraft to achieve common goals.
\end{enumerate}

The focal point of our paper is to identify a solution to have autonomous mission planning software onboard that is capable of making decisions, reacting to the external environment, and dealing with anomalies while keeping operability from ground.
Mission planning aims to allocate several tasks considering high-level goals, constraints, and resources. 
In recent years, researchers started studying this topic from different angles: using fuzzy neural networks to respond to uncertainty and proposing rescheduling \citep{LI2014678}, exploiting symmetric neural networks to improve heuristic search \citep{sym11111373}, using tabular search algorithms \cite{5608821}. All these methods fall into the category of \emph{Static Scheduling} because of the time they require to gather all the tasks.

To tackle this time issue, \emph{Dynamic Scheduling} builds a mathematical model considering five basic objects: tasks, resources, available opportunities, constraints, and objectives.
In this framework, scientists have been using novel heuristic algorithms to get \emph{close-to-optimal} results \cite{WANG2015110}; reinforcement learning based on neural network and transfer learning to solve decentralized Markov decision processes \citep{WANG2011493}; casting the problem as a stochastic dynamic Knapsack problem, and using deep reinforcement learning techniques to address it \citep{WANG20191011}. 
A downside of these approaches is that they require a large training dataset to output a robust decision-maker. Moreover, they often need frequent maintenance activities to be kept in line with the dynamic operational environment of a satellite. 

Other proposed solutions consist of hybrid procedures, mixing autonomous onboard capabilities and ground operations. 
A concrete example of these kinds of systems is the FireBird mission whose scheduler exploits very simple algorithms maintaining operability from ground \citep{lenzen2014onboard}; another one is The MER's scheduler that uses the concept of ``onboard execution of goal-oriented mission operations'' in addition to minor support from a ground process \citep{https://doi.org/10.1002/rob.20192}. 

For an \emph{Earth Observation} (EO) Satellite, all the spacecraft activities are scheduled by the Mission-Planning Engineer weekly. 
Then, the tasks are uploaded to the satellite as commands during ground visibility. 
New satellites can accept three types of commands:
	\begin{itemize}
		\item \emph{Multi Repeat Cycle} (MRC) commands, which are executed at the same orbit positions over the cycles.
		\item \emph{Single Repeat Cycle} (SRC) commands, which are executed at a specific orbit position for only one cycle.
		\item \emph{Mission Time Line} (MTL) commands, which fire at a specific time.
	\end{itemize}

In this paper, we present an algorithm capable of deciding when is the most suitable time to schedule the specific request to dump data from the onboard memory of a satellite (during visibility) over a ground station while keeping the system as simple as possible and  \emph{with no help from the ground}. 
We do this by casting our online decision-making problem into a prediction with expert advice setting \citep[Section 2]{cesa2006prediction} for which we implement a \emph{Follow-The-Leader} strategy \cite{kotlowski2018minimax} targeting SRC commands.

The paper is organized as follows. 
In \Cref{s:problem}, we give a high-level presentation of the specifics of the problem. 
In \Cref{s:setting}, we provide a formal mathematical model for it. 
In \Cref{s:ftl}, we introduce the Follow-The-Leader (FTL) strategy and discuss the theoretical motivations of this choice.
In \Cref{s:experiments}, we present the results of our experiments on real-life data, showing that implementing a simple FTL algorithm increases the performances of the satellites by more than $60\%$.

\section{The problem of optimizing Acquisition and Loss of Signal offsets for spacecraft memory dump}
\label{s:problem}
	EO Satellites are designed to perform remote sensing. Most of them operate at \emph{Low Earth Orbit} (LEO) with a period of approximately $120$ minutes. 
	Depending on the \emph{swath} of the onboard primary mission instrument, which is the device's \emph{areas} imaged on the surface, the satellite will complete a full Earth's coverage after a certain number of orbits. 
	This means the spacecraft completed a cycle. 
	Each separate orbit of a cycle is referred to as \emph{Relative Orbit} and is repeated through different cycles. 
	Motivation-wise, EO satellites play a fundamental role in weather forecast and natural disaster response scenarios, to name a couple.
	
	Spacecraft Operations Centers are in charge of distributing information to the users by commanding the satellites to perform activities over specific portions of the globe. This is done through the \emph{Mission Control Center} (MCC), which is capable of receiving telemetry and sending commands via the \emph{Mission Control System} software.
	
	Customers define the requirements and, or constraints that drive the operations performed during the lifetime of a mission.
	In particular, data \emph{timeliness} is a crucial parameter for the users' community: it is defined as the difference between the beginning of the sensing time of an instrument and the reception time of the same data on ground.
	 
	 Depending on the quality level of the products, data timeliness is categorized with different labels.
	 For Earth Observation satellites we can divide the delivered data into three main macro-categories:
	\begin{itemize}
		\item \emph{Near Real Time} (NRT), whose timeliness is lower than, or equal to the orbit period.
		\item \emph{Short Time Critical} (STC), whose timeliness can be hours.
		\item \emph{Non Time Critical} (NTC), whose timeliness can be days even weeks.
	\end{itemize}
	In this work, we focus on NRT data.
	To respect its requirements, NRT data must be dumped at every orbit without losing telemetry frames.
	Failing to do so results in critical data corruption, requiring another dump on the following orbit, therefore exceeding the timeliness. 
	To protect against data corruption, a Mission Planning Engineer uploads to the onboard computer software two commands sequences for each orbit: start and stop the memory dump during visibility time. 
	The timing of the two sequences relies on Flight Dynamic prediction of the passes over an antenna that defines the predicted \emph{Acquisition of Signal} (AOS) and \emph{Loss of Signal} (LOS).
	There are three notable categories for these pairs of events: AOS0 (resp., LOS0) is when the event occurs at the horizon, with an elevation of 0 degrees; 
	AOSM (resp., LOSM) denotes the event taking into account the masking of other objects, such as mountains, trees, or other antennas, therefore happening after AOS0 (resp., before LOS0); 
	AOS5 (resp., LOS5) is when the event takes place 5 degrees above the horizon, therefore it can happen before or after the masking event, depending on the specific conditions.
	
\begin{figure}
\centering
		
\begin{tikzpicture}[scale=0.95]
\draw[red] (50:5cm) -- (0,1.7) -- (125:5cm);
\draw[green] (45:5cm) -- (0,1.7) -- (135:5cm);
\draw ([shift=(45:5cm)]0,0) arc (45:135:5cm);
\draw[green] (45:5cm) node[right] {$LOS$}
             (135:5cm) node[left] {$AOS$};
\draw[red] (50:5cm) node[above right] {$L_t$}
             (125:5cm) node[above left] {$A_t$};
\draw[xshift=2cm, yshift=5.5cm, rotate=-30, scale=0.75, fill=blue!40!white]
    (0,0.3) -- (-1,0.2) -- (-1,0.8) -- (0,0.7)
    (0.5,0.3) -- (1.5,0.2) -- (1.5,0.8) -- (0.5,0.7)
    (0,0) rectangle (0.5,1)
;
\draw[xshift=2cm, yshift=5.5cm, rotate=-30, scale=0.75, gray, thick]
    ([shift=(250:0.5cm)]0.25,0) arc (250:290:0.5cm)
    ([shift=(245:0.5cm)]0.25,-0.1) arc (245:295:0.5cm)
    ([shift=(240:0.5cm)]0.25,-0.2) arc (240:300:0.5cm)
;
\draw[xshift=2cm, yshift=5.5cm, rotate=-30, scale=0.75, fill=blue!40!white]
    (0,-0.251) -- (0.5,-0.251) --
    ([shift=(0:0.25cm)]0.25,-0.25) arc (0:180:0.25cm) -- cycle;
\draw[fill=red!40!yellow, draw=black] (-0.5,0.4) -- (-0.1,1.1) -- (0.1,1.1) -- (0.5,0.4) -- cycle;
\draw[fill=red!40!yellow, draw=black] (-0.5,0.4) rectangle (0.5,0.5);
\draw[fill=red!40!yellow, draw=black] (0.5,1.5) -- (-0.5,1.5) -- ([shift=(180:0.5cm)]0,1.5) arc (180:360:0.5cm) -- cycle;
\draw[fill=red!40!yellow, draw=black] (-0.02,1.5) -- (0,1.7) -- (0.02,1.5) -- cycle;
\end{tikzpicture}
\caption{Pass' events timeline: AOS (resp., LOS) represents the highest events between AOS5 and AOSM (resp., LOS5 and LOSM); $A_t$ and $L_t$ represent the start and the stop of the dump which could be (and usually are) asymmetric with respect to the overall visibility.}
\end{figure}
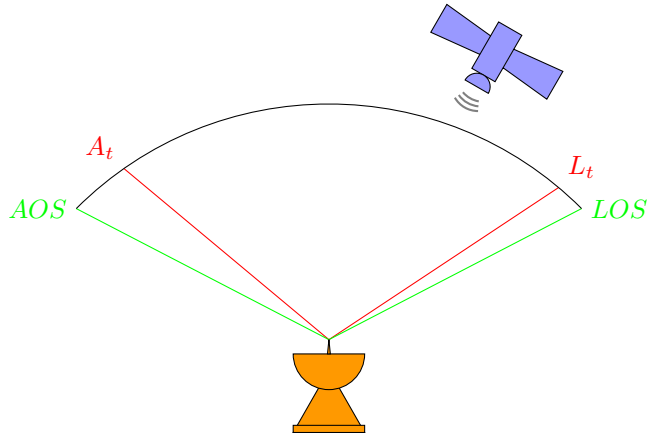

	In Spacecraft operations, the start of the dump is always shifted by a fixed amount of time, or \emph{offset}, with respect to the latest event between AOS5 and AOSM. 
	The same happens for the end of the dump which is anticipated with respect to the earliest between LOS5 and LOSM.
	This is done to guarantee the desired level of robustness against failures.

	These offsets are typically set manually and can be changed for each orbit or cycle. Usually, they are initialized with default values that consider the size of the dump and possible errors in the orbit determination and are then kept fixed, but during the early phase of a mission, things can diverge from the preparation phase, and offsets often need to be updated.
	In case of issues on the ground side (late acquisition and, or early loss of signal), data will be corrupted either because the satellite started dumping before locking over the antenna's signal or because the onboard computer was still downloading data while suddenly connection was cut off. 
	The only way to prevent loss of data is to modify the offsets \emph{ad-hoc} for each pass. 
	This operation is not currently carried on online automatically. Therefore, to recover after an issue of this type, engineers need to manually send a request through the \emph{Mission Control System} (MCS) at the next visibility opportunity. 
	Changing these values to avoid failure is an error-prone task that puts unneeded pressure on engineers. 
	There is no benefit in making these predictions manually.
	On the contrary, the nature of this problem makes it a perfect candidate for installing machine learning algorithms on onboard computers so that they can automatically learn from experience and predict online how to optimally set these offsets in order to optimize the data exchange with ground stations.

	In this paper, we propose an algorithm that selects the best AOL/LOS offsets at each cycle for every orbit. Moreover, we show that implementing this approach would not only avoid the need for operators to intervene manually, but it would save a vastly larger amount of the passes when compared to the currently employed \emph{ad-hoc} techniques.

\section{Formal Mathematical Setting}
\label{s:setting}

In this section we give a brief formal introduction to the online machine learning protocol that we will use to model our offset optimization problem.

We cast our repeated interactions with the environment in the \emph{prediction with expert advice} framework.
In this formulation, each problem instance is characterized by:
\begin{enumerate}
	\item A finite set of nonnegative times $\cOa$ 
    ---the offsets from the times of Acquisition Of Signal.
	\item A finite set of nonnegative times $\cOl$ 
    ---the offsets from the times of Loss Of Signal.
	\item A family $(p_{a,l})_{a \in \cOa, \, l \in \cOl}$ of numbers in $[0,1]$ 
	---the probabilities of a successful dump when some AOS and LOS offsets are selected.
\end{enumerate}
Only the \emph{action} set $\cOa\times\cOl$ is known to the learner and the best $p_{a,l}$ must be learned through data-collection.
The online protocol proceeds as in Online Protocol \ref{a:protocol}.
{
\makeatletter
\renewcommand{\ALG@name}{Online Protocol}
\makeatother

\begin{algorithm}[tb]
   \caption{}
   \label{a:protocol}
\begin{algorithmic}
    \FOR
    {%
    each time step $t = 1, 2, \dots$
    }
   \STATE The learner selects a pair of offsets 
   $(A_t,L_t) \in \cOa\times\cOl$
   \STATE The environment draws samples $B_t(a,l)\in\{0,1\}$ according to Bernoulli distributions with biases $p_{a,l}$, independently of each other and the history so far, for each pair of offsets $(a,l) \in \cOa\times\cOl$\;
   \STATE The learner gains the \emph{reward} 
   $B_t(A_t,L_t) \in \{0,1\}$
   \STATE The learner observes as \emph{feedback} the samples 
   $B_t(a,l)$, for all $(a,l)\in \cOa\times\cOl$
   \ENDFOR
\end{algorithmic}
\end{algorithm}
}

The objective of the learner is to collect as much reward as possible in any given time \emph{horizon} $T$.
More precisely, denoting by $\alpha$ the algorithm that generates the sequence of actions $(A_1,L_1)$, $\dots,$ $(A_T,L_T)$, the goal is to minimize the \emph{regret}
\begin{align*}
	R_T{(\alpha)}
&
\coloneqq
    \max_{(a,l) \in \cOa\times\cOl} \E \lsb{ \sum_{t=1}^T B_t(a,l) - \sum_{t=1}^T B_t(A_t,L_t) }
\\
&
=
    T \max_{(a,l)\in \cOa\times\cOl} p_{a,l}
    - \E\lsb{ \sum_{t=1}^T B_t(A_t,L_t) } \;.
\end{align*}

In words, we want the total reward accumulated by the learner to be as close as possible to that of an agents with full knowledge of the family of probabilities $(p_{a,l})_{(a,l)\in\cOa\times\cOl}$ that operates optimally by always choosing an action $(a^\star,l^\star)$ that maximizes its expected reward $p_{a^\star,l^\star}$.

\section{\ftl{}}
\label{s:ftl}

In this section, we present the \ftl{} paradigm (\Cref{a:ftl}). 
Our choice of applying this strategy to the offset-optimiza\-tion problem was initially motivated by compelling theoretical considerations (see \Cref{s:ftl-theory}).
We then confirmed these theoretical findings empirically, showing that an implementation of a \ftl{} algorithm does lead to dramatic performance improvements when applied to real-life scenarios (\Cref{s:experiments}).

\begin{algorithm}[tb]
    \caption{Follow the Leader (FTL)}
    \label{a:ftl}
\begin{algorithmic}
    \STATE {\bfseries Input:} action set $\cOa\times\cOl$, tie-breaking rule $\tau$
    \FOR
    {%
    each time step $t = 1, 2, \dots$
    }
   \STATE select action 
    \[
        (A_t,L_t) 
    \coloneqq
        \argmax_{(a,l) \in \cOa\times\cOl} \sum_{s=1}^{t-1} B_s(a,l) \;,
    \]
    breaking ties according to $\tau$, with the convention that (on time step $t=1$) $\sum_{s=1}^0 B_s(a,l) \coloneqq 0$, for all $(a,l)\in \cOa\times\cOl$\;
   \ENDFOR
\end{algorithmic}
\end{algorithm}

In words, a \ftl{} strategy maintains, during time steps $t$, an integer $\sum_{s=1}^{t-1} B_s(a,l)$ for each action $(a,l)$, representing how well the pair of offsets $(a,l)$ performed so far.
It then simply selects an action with the best past performance, breaking ties according to the rule $\tau$ that is passed to the algorithm as an input.
We highlight the role of the tie-breaking rule $\tau$ for two main reasons that are unfortunately often neglected.
Firstly, from a theoretical standpoint, ties have to be broken in a measurable way.\footnote{Although we will not insist excessively on the mathematical formalism, it is important to know that unexpected pathologies may occur if measurability is not guaranteed. In fact, the regret might not even be well-defined without it! For more on this topic, see, e.g., \cite[Section 2.4]{cesari2021nearest}.}
Secondly, depending on the application, some types of tie-breaking rules are typically preferred to others.
For example, in theoretical results, ties are typically broken uniformly at random (for mathematical convenience) but in practice, this is often unacceptable (because no practitioner would change a strategy that is currently working if no changes in the system are detected).
We will elaborate more on this key point in \Cref{s:experiments}.

\subsection{Theoretical Motivations}
\label{s:ftl-theory}

In this section, we give a concise summary of the theoretical results that motivated our choice of adopting a \ftl{} strategy for the offset optimization problem.

A recent paper by \citet[Corollary~4]{kotlowski2018minimax} shows that such a policy is optimal in the following strong sense.
 
\begin{theorem}
For any time horizon $T$ and any other online learning algorithm $\alpha$, the FTL algorithm with ties broken uniformly at random satisfies
\[
    \sup_{p} R_T(\mathrm{FTL})
\le
    \sup_{p} R_T(\alpha) 
\]
where the supremum is over all possible choices of probabilities $p_{a,l}\in[0,1]$, for $(a,l) \in \cOa\times\cOl$.
\end{theorem}

In words, the previous result states that no algorithm can achieve better performances than FTL in general.
\citet[Corollary~4]{kotlowski2018minimax} goes even further, showing that in addition to the regret, the same optimality holds for expected redundancy and excess risk.

Quantitatively, it is immediate to show that FTL enjoys a \emph{finite} regret, even when the time horizon $T$ approaches infinity, assuming that there is at least an action $(a,l)$ with $p_{a,l}=1$.
Indeed, in this case, for any $T$, 
\[
    R_T(\mathrm{FTL)} 
\le
    1 + \babs{ (a,l) \in \cOa\times\cOl : \mu_{a,l}\nu_{a,l} \in (0,1) } \;.
\]
The expression on the right-hand side counts the total number of mistakes that FTL could make: on round $1$, it could accidentally select an action $(a,l)$ with expected reward zero, i.e., a pair of offsets for which it is impossible to dump successfully. 
All these ``bad'' actions will be automatically excluded in all future rounds by definition of FTL.
Then, the only way the algorithm could pay some regret on a round $t \ge 2$ is if it selected an arm $(A_t,L_t)$ with expected regret $p_{A_t,L_t}<1$ which happened to have given reward $1$ in \emph{all} previous rounds but returned $0$ on the current one.
Again, by definition of FTL, such an action could only ever be played once and by the nature of the problem, the probability of this happening decreases exponentially with time.

These considerations give a strong theoretical foundation to the credence that FTL strategies outperform all other algorithms designed for this problem.
In the following section, we will show on real data that this is indeed the case.

\section{Implementation and Experimental Results}
\label{s:experiments}

The algorithm has been evaluated by comparing the results of a schedule produced with the autonomous choice of the offsets against a real operating schedule and the real telemetry data gathered from satellites.
	\subsection{System Configuration}
	The \emph{Mission Planning} (MP) software is a Commercial-Off-The-Shelf (COTS) written in C/C++ and it produces the schedule according to the schedule format from CSSDS standards \citep{schedule2018}. The software for autonomous selection of dump sequences has been written with Python v3.7. 
	We chose Python for the numerous support packages and modules which allow every user to personalize the tool with their own widgets, depending on their unique needs. 
	It is also simpler than the commonly used C++ and Java, and its lower code complexity makes it easy for beginners to approach it. 
	Simulations have been performed on a Windows 10 Pro x64 machine, processor i5-8250U 1.60 GHz, 8.00 GHz of RAM.
	
	The real telemetry data is provided by using SCOS-2000, a generic MCS software developed by the European Space Agency (ESA). 
	It provides the means for satellite operators to monitor and control one or more satellites. 
	The MCS is connected to the Antenna that the satellite is downloading over, and as soon as the spacecraft and antenna signals get \emph{on lock}, live telemetry is received and SCOS displays the first timestamp of the first telemetry frame received, doing the same when the pass is over and the last frame's timestamp is displayed. 
	These two pieces of information are then saved in a \emph{csv} file together with the relative orbit number and the cycle and passed to MP software. 
	This task has been performed by Spacecraft Controllers who were recording the time of the first and last TM frame ---but not regularly, leading to a partially filled set of data that required some pre-processing.
	All the test data set: telemetry, flight dynamic events, and schedule, are kind courtesy of Eumetsat.
	These data recordings begin at cycle 6  of the mission because in the first cycles 
	the satellite was not on the reference orbit, thus revolutions were not cyclic.
	
	Regarding the flight dynamic data, these are generated by a COTS software that takes telemetry data as input and generates a file, according to CCSDS standards \citep{orbitmsg2009}, which contains the propagation of the orbit of the satellite and its events (e.g. when the spacecraft is flying over a certain place at a certain time it save it as an event). 
	The file is then passed to the Mission Planning software as well.
	After receiving all these files, the Mission Planning engineer can generate a schedule which is an \emph{xml} file that is then sent back to MCS to generate the commands.
	\subsection{Scenario Configuration}
	The satellite considered for our tests is Copernicus Sentinel-6 and belongs to the European Earth Monitoring program. 
	It works in an LEO orbit with an altitude of $1336$ km, inclination $i$ of $66$ deg, and 127 orbits per cycle \citep{enwiki:1057089658}. 
	Each cycle lasts about 9.9 days. The average data dump is of 14 minutes. 
	A default pair of values of offsets for both AOS and LOS is evaluated according to the memory dump size; in this case, we started with $A_1=30s$ and $L_1=10s$.

	\paragraph{Tie-breaking rule}
	We break ties among leading AOS/LOS offsets by optimizing for both robustness and maximal usage of Station visibility.
	To do so, we first compute the smallest possible AOS and LOS offsets that could guarantee a successful dump given the historic data.
	Instead of selecting these offsets directly ---this could be prone to failures in early rounds--- we then compute a \emph{safe} pair of offsets, that would guarantee a successful dump that starts and ends at a maximal distance from the smallest possible AOS and LOS offsets computed above.
	
	\subsection{Baselines}
	The FTL algorithm has been compared against data coming from real telemetry: an initial schedule has been generated with offsets set as per an operational scenario, then the schedule has been uploaded to the satellite. 
	All the issues that occurred during the mission have been recorded together with the start and stop time of the real telemetry. This information has been compared against our FTL software.
	
	\subsection{Performances} 
\begin{figure}[ht!]
	    \centering
	    \begin{tikzpicture}[scale=0.8]
            \begin{axis} [ybar, height=7cm, width=11cm, xmin=0, xmax=7,ymin=0,ymax=42]
                \addplot coordinates {
                    (1,30) 
                    (2,30) 
                    (3,0) 
                    (4,30) 
                    (5,30) 
                    (6,30)
                };
                \addplot coordinates {
                    (1,13) 
                    (2,13) 
                    (3,0) 
                    (4,16) 
                    (5,16) 
                    (6,16)
                };
                \legend {AOS Offset [s], LOS Offset [s]};
            \end{axis}
            \node[above,font=\large] at (current bounding box.north) {RON 125};
        \end{tikzpicture}
        \caption{The LOS offset for orbit $125$, initialized as $10s$, is updated to $13s$ after time step 1, and again to $16s$ after time step 4. 
        Overall, only $1$ pass was lost after the initialization step. 
        As a reference, all $5$ recorded passes were lost in the baseline mission  (time step 3 was not recorded).}
        \label{fig:ron125}
	\end{figure}
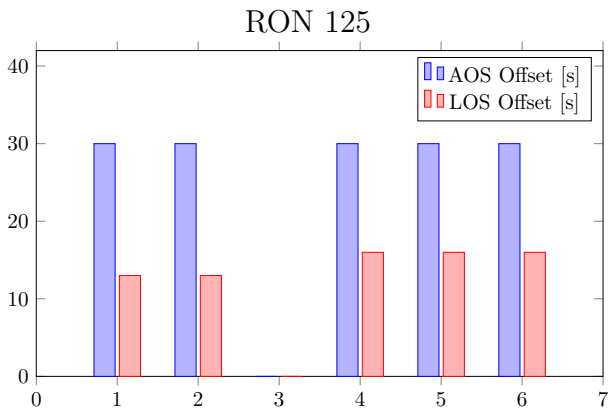

	The algorithm was run for 6 cycles of 127 orbits each. 
	During the first 6 life cycles of the baseline mission, a total of 762 passes occurred, and 67 of them contained corrupted data due to late/early AOS/LOS. 
	When this happens, passes need to be re-dumped the next orbit, losing timeliness. 
	With our FTL algorithm, we were able to save more than $62\%$ of the corrupted passes, leading to a dramatic improvement in timeliness. 
	This was done without the need for any manual intervention. 
	The few passes that were still lost  occurred ad the very early stages of the process, where the choices of \emph{any} decision-maker are doomed by the variance.
	For a pictorial example of the behavior of FTL,  see \Cref{fig:ron125}.
	
\section{Conclusions}

We presented a lightweight algorithm capable of updating the spacecraft command sequences online, autonomously, and considerably improving user requirements for data timeliness.
More precisely, we manage to reach two competing goalposts simultaneously. First, \emph{simplicity}: the Python implementation we provide consists of a relatively minute number of instructions, and intrinsically simple routines are quick to explain to engineers used to different system practices while requiring effectively no maintenance. 
Second, \emph{high performances}: our algorithm vastly outperforms the existing \emph{ad hoc} methods, as our testing showed that switching to our FTL routine would save an enormous amount of otherwise lost data.

\section*{Acknowledgements}
The authors gratefully acknowledge the support of the Eumetsat Mission Planning team where Jonathan is working.
This work started during Tommaso's Post-Doc at the Institut de Mathématiques de Toulouse, France, and benefited from the support of the project BOLD from the French
national research agency (ANR).
Tommaso Cesari also gratefully acknowledges the support of IBM.

\bibliographystyle{plainnat}
\bibliography{biblio}


\end{document}